\documentclass{article}

\usepackage{arxiv}

\usepackage[utf8]{inputenc} % allow utf-8 input
\usepackage[T1]{fontenc}    % use 8-bit T1 fonts
\usepackage{hyperref}       % hyperlinks
\usepackage{url}            % simple URL typesetting
\usepackage{booktabs}       % professional-quality tables
\usepackage{amsfonts}       % blackboard math symbols
\usepackage{nicefrac}       % compact symbols for 1/2, etc.
\usepackage{microtype}      % microtypography
\usepackage{lipsum}
\usepackage{graphicx, multirow}
\usepackage{amsmath}
\usepackage{algorithm} 
\usepackage{algpseudocode} 
\AtBeginDocument{%
  \providecommand\BibTeX{{%
    \normalfont B\kern-0.5em{\scshape i\kern-0.25em b}\kern-0.8em\TeX}}}

\title{Multiple Imputation with Denoising Autoencoder using Metamorphic Truth and Imputation Feedback}

\author{
  Haw-minn Lu \\
  Gary and Mary West Health Institute\\
  San Diego, CA 92037 \\
  \texttt{hlu@westhealth.org} \\
   \And
  Giancarlo Perrone \\
  Gary and Mary West Health Institute\\
  San Diego, CA 92037 \\
  \texttt{gperrone@westhealth.org} \\
  \And
 Jos\'{e} Unpingco \\
  Gary and Mary West Health Institute\\
  San Diego, CA 92037 \\
  \texttt{jhunpingco@westhealth.org} 
}

\begin{document}
\maketitle
\begin{abstract}
Although data may be abundant, complete data is less so, due to missing  columns or rows. This missingness undermines the performance of downstream data products that either omit incomplete cases or create derived completed data for subsequent processing. Appropriately managing missing data is required in order to fully exploit and correctly use data. We propose a Multiple Imputation model using Denoising Autoencoders to learn the internal representation of data. Furthermore, we use the novel mechanisms of Metamorphic Truth and Imputation Feedback to maintain statistical integrity of attributes and eliminate bias in the learning process. Our approach explores the effects of imputation on various missingness mechanisms and patterns of missing data, outperforming other methods in many standard test cases. 

\end{abstract}

\section{Introduction}
Missing data is problematic, ubiquitous, and lacks a universal solution. Most often, incomplete observations are simply omitted, thus shrinking the dataset, along with any potential gains from that data. One common approach is to simply replace the missing values with the mean/median of the observed values in the same partition. Other approaches use a machine learning model to preprocess and fill-in the missing observations. While these methods superficially resolve the missing data, they may also lead to downstream bias and unexpected errors. 

Why data is missing is a further difficult problem, requiring domain knowledge and understanding of the data collection process. Did a corrupted input lead to an empty response in an online survey? Or was that field intentionally skipped? Understanding the mechanism of missing data drives the effectiveness of the imputation method \cite{RUBIN}. Imputing missing observations by the mean/median fails to account for the uncertainty of missing values during the imputation procedure. To circumvent this, the process of imputing may be repeated, producing multiple imputed values that are scholastically different. This is known as \emph{Multiple Imputation} \cite{doi:10.1177/096228029900800102}, which involves repeatedly performing imputation inferences on any missing data, and then combining the results to analyze imputed results incorporating any previous uncertainty for missing data. This procedure recognizes the uncertainty of predictions by introducing variability \cite{Sterneb2393} into imputed values.

In section~\ref{sec:background}, we described the terminology and formal concepts relating to missingness as well as the previous use of autoencoders in multiple imputation. In section~\ref{sec:architecture}, we describe the architecture of the underlying deep neural network used. In section~\ref{sec:method}, we describe the training method. A catalog of data used is described in section~\ref{sec:data}. The experiment is detailed in section~\ref{sec:experiments}. The results are presented in section~\ref{sec:results}. Section~\ref{sec:conclusion} discusses the implications of the results.

\section{Background and Related Work}\label{sec:background}

\subsection{Missingness Models}

Consistent with recent literature, we distinguish between missingness \textit{mechanisms} and missingness \textit{patterns}. Introduced by Rubin, \cite{RUBIN} missingness mechanisms refers to one of three  situations: missing completely at random (MCAR), missing at random (MAR) and missing not at random (MNAR). Additionally, we adopt the table nomenclature for a dataset where \textit{rows} are synonymous with observations, \textit{columns} with attributes and \textit{cells} with attributes of an observation. If a particular column cannot have any missing rows (i.e., not subject to missing data), that column is \textit{permanent}. Otherwise, it is referred to as \textit{vulnerable}.

\begin{itemize}
\item MCAR -- A row has one or more missing cells independent of the values of the cells in that row.

\item MAR -- Unlike MCAR, a missing cell is a probabilistic function of the permanent columns. For example, the local school district surveys families regarding household income, while knowing which families participate in the school free lunch program.  Families who participate in the program may be less likely to respond to the household income question. Thus, the column indicating participation in the free lunch program is \textit{permanent} but the household income data may be missing for a particular row based on program participation.  

\item MNAR -- All cases other than MCAR and MAR. This implies that the missingness is a function of variables in all columns, permanent or not. Thus, a cell could be eliminated probabilistically based upon its own value, which means that it does not appear in the resulting dataset. For example, an overweight respondent might be less likely to report his/her body weight so the resulting data set may not contain a value for body weight for some rows. 
\end{itemize}

\noindent We refer to missingness \textit{patterns} as random and uniform.  A \textit{uniform} missingness pattern means that all vulnerable columns in a particular row are missing. A \textit{random} pattern means that some elements in the vulnerable columns are missing probabilistically. In summary, the missingness mechanism decides whether a  particular row has missing data and the missingness pattern indicates which cells in a given row are subject to missingness.  

\subsection{Autoencoders}

An autoencoder in the context of artificial neural networks and in particular deep neural networks, is a neural network that is trained to copy the input as the output (i.e, approximate the identity function). Rumelhart, {\it et. al.}\cite{Rumelhart} showed that they can be used to learn a hidden internal representation of the data. Denoising Autoencoders (DAE)\cite{10.1145/1390156.1390294} are designed to remove noise from data. They are characterized by high dimension in the hidden layers and with stochastic corruption (dropout) at the input. DAEs are employed by Gondara\cite{DAE} to perform multiple imputation. The research presented here is based on the deep neural network employed by Gondara and is explained in greater detail below. There are additional methods employing autoencoders to attack the problem of imputation\cite{Tran_2017_CVPR}. 

\section{Architecture}\label{sec:architecture}

\begin{figure}[ht]
  \centering
  \includegraphics[width=180pt]{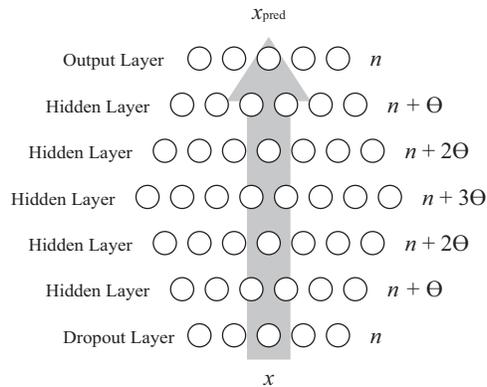}
  \caption{Denoising Autoencoder}
  \label{fig:dae}
\end{figure}

Our autoencoder architecture is essentially identical to that used by Gondara. The main difference lies in the training method described in the next section. Our DAE architecture employs most of the same hyperparameters as Gondara. As shown in Figure~\ref{fig:dae}, the first layer implements a dropout (stochastic corruption) of 50\%, there are 5 additional hidden layers with the layers first increasing in width by the hyperparameter $\Theta$ from the previous layer then decreasing by the same $\Theta$ in the final layers, where $\Theta=7$. All of these layers, except for the output layer, use a $\tanh$ activation function. The depth of the network, the value of $\Theta$,  the activation function and degree of dropout are as recommended by Gondara. Additionally, the DAE architecture is implemented using Tensorflow and using the Adam optimizer\cite{DBLP:journals/corr/KingmaB14}.

Other factors which can be treated as hyperparameters include the initial treatment of missing data and the proportion of data used in training. Because training neural networks requires all values to be present, missing values must be imputed with some value. A more detailed discussion of initial imputation methods is given in the next section. Additionally, the proportion of data used in training is more appropriately discussed in the next section.

\section{Method}\label{sec:method}

The method of training the DAE is significantly different than that of training a DAE or that of a typical deep neural network. First, we discuss the motivation behind our training method and then the method of metamorphic truth and feedback. Finally, we discuss the preparing the training data.

\subsection{Motivation}

Deep neural networks are not designed to accommodate missing data. Therefore, imputation techniques using deep neural networks perform some type of initial imputation to fill in missing data in a training set prior to training. For example, Gondara initially imputes the data using mean-imputation for the numerical columns on an architecture identical to ours. However, the initial imputation is extremely important for downstream performance of the DAE. For example, Table~\ref{tab:imp} shows imputing the Boston Housing Data using the DAE under MAR mechanism with random pattern where initial imputation is performed five times. The mean over these imputations is shown with the corresponding maximum value over those imputations in parenthesis. The Max Imputed column uses the maximum value of the respective columns for the initial imputation, intuitively a poor choice. The Perfect Guess column fills in the true value of the missing cell for the initial imputation, an obviously unrealistic best case initial imputation. This table shows that maximum initial imputation has the worst 
$\mathrm{RMSE_{sum}}$, which is approximately twice that of a mean-value initial imputation. Note that the Perfect Guess initial imputation is half that of the mean imputed initial imputation. This table shows that the initial imputation severely affects downstream performance of the DAE. 

In this example as well as throughout our experiments $\mathrm{RMSE_{sum}}$ is used to measure the quality of a given imputation. The $\mathrm{RMSE_{sum}}$ is defined by equation~\ref{eq:rmse}.

\begin{equation}
\label{eq:rmse}
\mathrm{RMSE_{sum}} = \sum_{j=1}^{N_{\mathrm{row}}}
\sqrt{\mathbb{E}\left(\sum_{i=1}^{N_{\mathrm{col}}} (\tilde{t}_i - t_i)^2\right) },
\end{equation}

\noindent where $N_{\mathrm{row}}$ are the number of rows, $N_{\mathrm{col}}$ are the number of columns, $\tilde{t}_i$ is an imputed cell, and $t_i$ is the corresponding cell from the original uncorrected row  and $\mathbb{E}$ is the expected value or mean. Note that only imputed cells appear in this equation.

There are two factors which contribute to this sensitivity to initial imputation. First, the initial imputed value is fed back to the neural network as the ground truth during training. This essentially tells the DAE that the \textit{correct} answer is the \textit{initial} imputed value, which biases the DAE towards the initial imputed value, and potentially away from a better value. As a practical matter, for many datasets,  the mean is  a good choice for imputation when considering \textit{mean} squared error. Because of this, if the mean is close to the good imputed value the feedback to the DAE will have less impact.

The second factor is that as the DAE produces better imputed values, the input and output diverge, which is \textit{contrary} to the learning objective for the DAE, which is to reproduce the input as the output, but without the noise.

\begin{table}
  \caption{Effect of Initial Imputation on DAE Imputation}
  \label{tab:imp}
  \centering
  \begin{tabular}{cccc}
    \toprule
    & Mean Imputed&Max Imputed&Perfect Guess\\
    \midrule
    $\mathrm{RMSE_{sum}}$   & 7.43 (7.59) & 16.72 (16.92) & 3.16 (3.32) \\
  \bottomrule
\end{tabular}
\end{table}

\subsection{Metamorphic Truth and Feedback}

Since an initial imputation must be performed to use the DAE, we use mean imputation. However, because our method mitigates the two factors contributing to initial imputation sensitivity mentioned above, the choice of the initial imputation is not as important. To mitigate the first factor, {\it metamorphic truth} is applied and to mitigate the second factor we feedback interim imputed values as the imputation for inputs to subsequent training epochs. In short, metamorphic truth decouples the DAE's output from the initial imputation and the feedback decouples the DAE's input from the initial imputation.

\begin{figure}[ht]
  \centering
  \includegraphics[width=240pt]{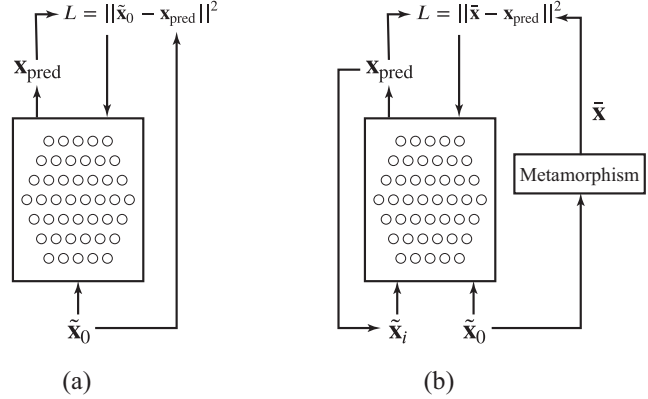}
  \caption{Denoising Autoencoder Learning}
  \label{fig:arch}
\end{figure}

Figure~\ref{fig:arch}a shows the typical training used traditionally with DAE in imputation. Given a row $\mathbf{x}$, missing values are imputed as $\tilde{\mathbf{x}}_0$. The initial imputed value $\tilde{\mathbf{x}}_0$ is fed into the DAE which produces a prediction $\mathbf{x}_\mathrm{pred}$ (we keep the notation of $\mathbf{x}$ rather than $\mathbf{y}$ as this is an autoencoder). A mean squared error (MSE) loss is computed by equation~\ref{eq:mse}

\begin{equation}
  \label{eq:mse}
  L = ||\tilde{\mathbf{x}}_0- \mathbf{x}_\mathrm{pred}||^2
\end{equation}

\noindent and then fed back into the DAE to be used to adjust weights by a proscribed optimization algorithm. Figure~\ref{fig:arch}b shows the modified learning process used to address the two aforementioned issues. To address the first concern mentioned above, we employ a technique called {\it metamorphic truth}. Metamorphic truth is a technique used to change the truth reported to the optimizer based on the prediction, but having the neural network treat the {\it truth metamorph} $\bar{x}$ as a constant. This distinguishes the technique from simply being a more complicated loss function. 

\noindent For imputation we use the following metamorphism, $\mathcal{M}$:

\begin{equation}
\mathcal{M}(\mathbf{x},\mathbf{x}_\mathrm{pred})_i =
\begin{cases}
{x_\mathrm{pred}}_i&\text{If $x_i$ is a missing attribute}\\
x_i&\text{Otherwise}\\
\end{cases}
\end{equation}

\noindent For the $i^{th}$ cell in a row, the truth metamorph is therefore expressed as

\begin{equation}
\bar{x}=\mathcal{M}(\tilde{\mathbf{x}}_0,\mathbf{x}_\mathrm{pred}).
\end{equation}

\noindent and the loss function can now be expressed as 

\begin{equation}
  L = ||\bar{\mathbf{x}}- \mathbf{x}_\mathrm{pred}||^2.
\end{equation}

We address the second aforementioned issue (divergence between input and output) by using the predicted values to impute subsequent training of the DAE. The new learning algorithm is given by Algorithm~\ref{alg:MYALG} which has two phases, (1) a priming phase and (2) a fine-tuning phase. In the priming phase, the DAE is trained for $N_\mathrm{prime}$ epochs on the initial imputed values (we initially impute with the mean). The purpose of this phase is to coarsely train the DAE, in principle 10-20 epochs is found to be sufficient for the datasets considered here. In the fine-tuning steps, the output of the DAE is used to compute a new imputation, and that imputed dataset is then used to train the DAE for $N_\mathrm{step}$ epochs. The new imputation only takes the predicted output for the \textit{missing}  attributes and leaves \textit{observed} attributes alone. This process is repeated for the desired number of steps. In our experiments, $N_\mathrm{step}=2$ though $N_\mathrm{step}=1$ can be used.

\begin{algorithm}
	\caption{Imputation} 
	\label{alg:MYALG}
	\begin{algorithmic}[1]
		\For {Number of imputations required}
		\State Set initial imputation $\tilde{\mathbf{x}}_0$
		\State Train DAE with $\tilde{\mathbf{x}}_0$ for $N_\mathrm{prime}$ epochs
		\State Run DAE on $\tilde{\mathbf{x}}_0$ to produce $\mathbf{x}_{\mathrm{pred}}$
			\For {$i=1\dots$ Number of steps required}
			    \State Impute $\tilde{\mathbf{x}}_i$ using $\mathbf{x}_{\mathrm{pred}}$
				\State Train DAE with $\tilde{\mathbf{x}}_i$ for $N_\mathrm{step}$ epochs
		        \State Run DAE on $\tilde{\mathbf{x}}_i$ to produce $\mathbf{x}_{\mathrm{pred}}$
			\EndFor
			\State take last $\tilde{\mathbf{x}}_i$ as the imputed dataset
		\EndFor
	\end{algorithmic} 
\end{algorithm}

\subsection{Training Methodology}

In a typical deep neural network training process, part of a dataset is held back for validation purposes. For example, Gondara separates datasets samples into approximately 70\% training set and 30\% test set. The main purpose of a validation data is to gauge when a neural network is starting to overfit. In short, the validation set is used to indicate when to stop training. In an {\it open learning} system, overfitting can be detrimental because the neural network is will not adapt to future unknown inputs. Hence validation data is used to measure "model accuracy," the accuracy applied to the entire open ended problem.  However, DAEs used for imputation are {\it closed learning} systems, that is once trained no future unknown inputs need to be accounted for. Hence, the detrimental effect of potential overfitting is not significant in a closed learning system and the model accuracy can be measured only on the training set since the training set represents the entire data set both present and future.

Beyond the use of a test set to detect overfitting in an open learning system, a test set is also used to evaluate the accuracy of a given deep neural network model. For the experiments presented here, true accuracy of the DAE is reflected in how well the imputed values compare to the original values that are removed as part of the experiment with measures such as $\mathrm{RMS_{SUM}}$ and covariance drift as defined below. Therefore, there is no need for a test set to be held in reserve to evaluate model accuracy as such model accuracy may not reflect the actual accuracy of the imputation.

Since there is no longer a compelling reason to reserve some samples for a test set, we elected to use {\bf all samples} for training. As a test set would not influence DAE training, omission of samples reserved in a test set would risk eliminating potential covariance relationships or other joint statistical relationships present in the omitted samples.

\section{Data}\label{sec:data}

Six datasets are used for the evaluation of the models are taken from \cite{Dua:2019} and \cite{StatLib}\footnote{obtained at http://lib.stat.cmu.edu/datasets/boston}. Five of the six are used by Gondara. Table~\ref{tab:ds} shows the name of the datasets with their abbreviations along with the number of columns and rows\footnote{It should be noted that the BC dataset had 699 rows which included 16 rows with missing cells. Those observations were dropped from the dataset.} in the dataset. For the same reasons as in the previously cited work, the motivation for the database selection is to demonstrate the effectiveness of the imputation model even under low-dimensionality or low sample size datasets. 

We adopt the convention set forth by Li\cite{Li2013} for inducing missingness in the datasets. For each dataset, columns are designated permanent (i.e., not subject to missingness) and vulnerable (i.e., subject to missingness). The specific designation of attributes for each category was arbitrary except that \textit{only} numeric and \textit{not} categorical columns were classified as vulnerable, in order to keep the experiment simple. However, categorical columns are retained as permanent columns so that they may potentially serve as covariates for imputation. 
Proper consideration of categorical attributes would entail exploring proper ways to encode those categories.

\begin{table}
  \caption{Datasets}
  \centering
  \label{tab:ds}
  \begin{tabular}{lrr}
    \toprule
    Name & Rows & Columns\\
    \midrule
Boston Housing (BH) & 506 & 14\\
Glass (GL) &214 &10\\
Ionosphere (IS) &351 &35\\
Breast Cancer (BC) &683 & 11\\
Sonar (SN) &208 &61\\
Wine (WN) &4898 &11 \\  
\bottomrule
\end{tabular}
\end{table}

\section{Experiments}\label{sec:experiments}
\subsection{Inducing Missingness}

Missingness is induced using the three mechanisms described above, MCAR, MAR and MNAR. In addition we apply a missingness pattern of either random or uniform. Our method of inducing the missingness involves two tuning probabilities $p_m$ and $p_p$, the former is used in the mechanism and the latter, the pattern. These are probabilities that can be adjusted to obtain the desired level of missingness and do not necessarily reflect on the probability an attribute is missing.

For MCAR missingness, a row has missing values with probability $p_m$. Clearly this is purely random and missingness is not a function of the values in the cells. Our construction of MAR and MNAR missingness is algorithmically the same, two random columns ($i,j$) are selected for the entire dataset. If for a given row $\mathbf{x}$ has $x_i>\mu_i$ and $x_j>\mu_j$ where $\mu_i$ and $\mu_j$ are the means of those two columns, then the observation has missing values with probability $p_m$. For MAR, the two selected columns are drawn from only the permanent columns and for MNAR they are drawn from the vulnerable columns. In the case of the uniform missingness pattern, if a row is induced by the mechanism to have missing data then the \textit{entire} row of the across all vulnerable columns is removed; otherwise, for the random missingness pattern,  the cells along that row are removed by the missingness mechanism with with probability $p_p$. 

For each dataset, a case of MCAR uniform, MCAR random, MAR uniform, MAR random, MNAR uniform and MNAR random are induced and saved so that the four imputation methods mentioned below can be applied to the same corrupted datasets. Generally our target for all cases is to induce a missing proportion between 14\% and 20\%\footnote{However in some cases, reaching that percentage proved challenging so the proportion of missing data for those cases we lowered the threshold to 10\%} achieved by adjusting the tuning probabilities. The actual percentage of cells with missing values are given in Table~\ref{tab:missing}.

\begin{table}
  \caption{Percentages of Cells Missing}
  \centering
  \label{tab:missing}
  \begin{tabular}{|c|c|c|c|c|c|c|}
    \hline
    & \multicolumn{3}{|c|}{Random} & \multicolumn{3}{|c|}{Uniform} \\
    \hline
    Dataset & MCAR & MAR & MNAR & MCAR & MAR & MNAR  \\ \hline
    BH& 15.8\%& 17.8\%& 15.2\%& 16.1\%& 17.0\%& 15.5\%\\ \hline
    GL& 14.4\%& 14.0\%& 14.2\%& 16.1\%& 14.2\%& 14.1\%\\ \hline
    IS& 20.4\%& 14.6\%& 10.0\%& 14.2\%& 12.1\%& 10.6\%\\ \hline
    BC& 16.4\%& 19.4\%& 18.1\%& 16.7\%& 15.1\%& 16.1\%\\ \hline
    SN& 16.0\%& 15.9\%& 10.2\%& 16.7\%& 14.7\%& 12.1\%\\ \hline
    WN& 16.1\%& 14.0\%& 11.8\%& 14.3\%& 10.2\%& 10.7\%\\ \hline

  \end{tabular}
\end{table}

\subsection{Comparison to Other Methods}

Comparative experiments are conducted comparing four imputation models. First is mean imputation where the mean of an attribute is substituted for the missing values. Second is Multivariate Imputation by
Chained Equations (MICE) \cite{MICE}, which is considered the state-of-the-art multiple imputation model. Third is the DAE presented by Gondara, which our model relies heavily upon. Fourth is the DAE with Metamorphic Truth and Imputation Feedback (DAE MT) presently described above.

\newcommand{\boxup}[1]{\begin{tabular}{@{}c@{}}#1\end{tabular}}
\begin{table}
  \caption{Comparison of imputation techniques ($\mathrm{{RMS}_{SUM}}$)}
  \centering
  \label{tab:result}
  \begin{tabular}{|c|c|c|c|c|c|c|c|c|c|c|}
  \hline
    & \multirow{2}{*}{\boxup{\rotatebox[origin=c]{90}{D.set}}}&\multicolumn{4}{|c|}{Random} & \multicolumn{4}{|c|}{Uniform} \\
    \cline{3-10}
     &  & DAE MT & DAE & MICE & Mean & DAE MT & DAE & MICE & Mean \\ \hline
    \hline
\multirow{6}{*}{\boxup{\rotatebox[origin=c]{90}{MCAR}}}
& BH & \textbf{6.7} (6.8) & 8.0 (8.1) & 8.4 (8.8) & 9.4 & \textbf{6.5} (6.6) & 7.8 (7.9) & 9.9 (10.3) & 8.4 \\ \cline{2-10}
& GL & \textbf{4.3} (4.4) & 4.9 (5.1) & 5.8 (6.8) & 5.7 & \textbf{4.4} (4.5) & 5.0 (5.1) & 7.3 (8.5) & 5.6 \\ \cline{2-10}
& IS & \textbf{17.7} (18.0) & 20.7 (20.9) & 23.1 (23.6) & 23.3 & \textbf{21.6} (21.7) & 23.9 (24.1) & 27.4 (28.4) & 25.9 \\ \cline{2-10}
& BC & \textbf{4.4} (4.5) & 5.6 (5.6) & 6.1 (6.3) & 6.7 & \textbf{5.3} (5.4) & 7.1 (7.2) & 6.6 (6.9) & 7.7 \\ \cline{2-10}
& SN & \textbf{28.6} (28.9) & 34.2 (34.3) & 37.3 (38.5) & 39.5 & \textbf{37.4} (37.5) & 39.1 (39.2) & 49.9 (50.6) & 39.7 \\ \cline{2-10}
& WN & \textbf{5.5} (5.6) & 6.2 (6.3) & 8.0 (8.1) & 6.9 & \textbf{6.2} (6.3) & 6.8 (6.8) & 8.8 (9.1) & 7.1 \\ \hline
\hline
\multirow{6}{*}{\boxup{\rotatebox[origin=c]{90}{MAR}}}
& BH & \textbf{5.4} (5.5) & 7.4 (7.5) & 7.9 (8.4) & 8.4 & \textbf{5.8} (5.9) & 8.3 (8.4) & 10.0 (10.3) & 8.8 \\ \cline{2-10}
& GL & \textbf{2.8} (2.9) & 3.5 (3.6) & 6.2 (7.6) & 4.1 & \textbf{7.4} (7.4) & 7.8 (7.8) & 10.2 (10.7) & 8.1 \\ \cline{2-10}
& IS & \textbf{19.0} (19.2) & 19.9 (20.0) & 25.5 (26.2) & 21.3 & \textbf{17.2} (17.3) & 19.4 (19.5) & 24.1 (24.7) & 21.4 \\ \cline{2-10}
& BC & \textbf{2.1} (2.4) & 4.4 (4.6) & 2.4 (2.8) & 5.4 & \textbf{3.2} (3.3) & 4.7 (4.8) & 4.3 (4.5) & 5.1 \\ \cline{2-10}
& SN & \textbf{35.5} (35.7) & 39.9 (40.2) & 50.8 (52.7) & 41.5 & \textbf{36.1} (36.3) & 38.9 (39.2) & 50.3 (51.6) & 39.4 \\ \cline{2-10}
& WN & \textbf{5.8} (5.8) & 6.3 (6.3) & 8.5 (8.5) & 6.7 & \textbf{5.9} (5.9) & 6.3 (6.3) & 8.6 (8.7) & 6.6 \\ \hline
\hline
\multirow{6}{*}{\boxup{\rotatebox[origin=c]{90}{MNAR}}}
& BH & \textbf{6.6} (6.7) & 8.4 (8.5) & 8.8 (8.9) & 9.4 & \textbf{9.4} (9.6) & 12.9 (13.1) & 12.2 (12.4) & 13.6 \\ \cline{2-10}
& GL & \textbf{3.1} (3.2) & 3.9 (3.9) & 5.6 (6.3) & 4.4 & \textbf{4.2} (4.3) & 5.1 (5.2) & 6.6 (7.2) & 5.6 \\ \cline{2-10}
& IS & \textbf{17.6} (17.7) & 20.7 (21.0) & 23.0 (23.6) & 23.3 & \textbf{21.5} (21.8) & 25.7 (26.1) & 27.4 (27.8) & 28.2 \\ \cline{2-10}
& BC & \textbf{2.2} (2.3) & 4.5 (4.6) & 2.6 (2.8) & 5.5 & \textbf{2.7} (2.9) & 4.8 (4.9) & 3.0 (3.3) & 5.3 \\ \cline{2-10}
& SN & \textbf{40.1} (40.4) & 46.1 (46.5) & 53.4 (55.1) & 49.4 & \textbf{34.0} (34.5) & 36.0 (36.2) & 48.2 (49.7) & 36.7 \\ \cline{2-10}
& WN & \textbf{5.6} (5.6) & 6.1 (6.1) & 8.5 (8.5) & 6.5 & \textbf{6.9} (7.0) & 7.7 (7.8) & 9.4 (9.5) & 8.1 \\ \hline
\end{tabular}
\end{table}

All measures applied to the quality of the imputations are computed on datasets that are normalized. For all imputation methods except MICE (MICE does not need it), data is normalized prior to the application of the model. However, in the case of MICE, the imputed output data is normalized using the same scaling factors used to normalize the other methods for fair comparison. Normalization is performed on each cell by subtracting the mean of that column and dividing by the standard deviation. 
The MICE program was run using default settings with 5 imputations. The DAE was trained for 500 epochs. The DAE MT was primed for 10 epochs ($N_\mathrm{prime}$) and 245 steps of 2 epochs ($N_\mathrm{step}$) each, totalling the same 500 epochs as the DAE. Both DAE MT and DAE were run 5 times for 5 imputations.
		 
\section{Results}\label{sec:results}
Table~\ref{tab:result} shows the results from running the various models on all the datasets. For each multiple imputation model, the mean and the maximum error ($\mathrm{RMS_{SUM}}$) across the imputations are shown with the maximum in parenthesis and the lowest value shown in bold. Since mean imputation is a single imputation, a single error is reported. From the table,  DAE MT outperforms both standard DAE, MICE and mean imputation in all cases studied. In many applications such as machine learning applications such as classification, minimizing $\mathrm{RMS_{SUM}}$ is very important, but for more traditional statistical analytics, the relationship between values in different columns is most important. 

As a rough measurement to the how well these relationships are preserved, we defined a {\it covariance drift}, which is basically the RMS average of the difference between the respective covariances of the original data and that of the data post-imputation. Mathematically, if the elements of the covariance matrix of the uncorrupted  data set is   $\sigma_{i,j}$ and the elements of the covariance matrix of the imputed data set is $\tilde{\sigma}_{i,j}$ then the drift is given by equation~\ref{eq:drift}.

\begin{equation}
  \label{eq:drift}
  \mathrm{drift_{cov}}=\frac{1}{N(N-1)}\sqrt{\sum_{i\ne j} (\sigma_{i,j}-\tilde{\sigma}_{i,j})^2}
\end{equation}

\noindent In the interest of space, Table~\ref{tab:cov} shows only the covariance drift for the MNAR missingness case for all the datasets. Once again, the mean and the maximum values (in parenthesis) of the covariance drift are shown with the lowest drift values in bold. The results are multiplied by 10 to be displayed in the table for easier comparison. With regard to the covariance drift, MICE generally outperforms both DAE models, however DAE MT outperforms significantly standard DAE in all cases and in some cases exceeds the performance of MICE. Hence, it is clear that the bias asserted by the initial imputation has a strong impact on the covariance which is significantly mitigated by DAE MT. 
Unsurprisingly, MICE does best in minimizing covariance drift in general as MICE strives to preserve these relationships whereas neural networks aim to minimize mean squared errors.

\begin{table}
  \caption{Covariance Drift under MNAR Missingness ($\times$10)}
  \centering
  \label{tab:cov}
  \begin{tabular}{|c|c|c|c|c|c|c|c|c|}
    \hline
    & \multicolumn{4}{|c|}{Random} & \multicolumn{4}{|c|}{Uniform} \\
    \hline
    Dataset & DAE MT & DAE & MICE & Mean & DAE MT & DAE & MICE & Mean \\
    \hline
BH & 0.6 (0.6) & 1.1 (1.2) & \textbf{0.4} (0.5) & 1.3 & 1.3 (1.5) & 2.6 (2.6) & \textbf{1.2} (1.3) & 2.7 \\ \hline
GL & \textbf{0.4} (0.4) & 0.6 (0.6) & 0.5 (0.6) & 0.7 & \textbf{0.7} (0.7) & 0.9 (1.0) & 0.8 (0.9) & 1.0 \\ \hline
IS & 0.6 (0.6) & 0.7 (0.7) & \textbf{0.5} (0.6) & 0.8 & 0.8 (0.8) & 1.1 (1.1) & \textbf{0.6} (0.7) & 1.2 \\ \hline
BC & 0.4 (0.5) & 1.1 (1.2) & \textbf{0.2} (0.2) & 1.4 & 0.4 (0.5) & 0.9 (0.9) & \textbf{0.1} (0.2) & 1.0 \\ \hline
SN & \textbf{0.5} (0.5) & 0.7 (0.7) & 0.6 (0.7) & 0.8 & \textbf{0.4} (0.4) & 0.5 (0.5) & 0.5 (0.5) & 0.5 \\ \hline
WN & \textbf{0.4} (0.4) & 0.5 (0.5) & 0.6 (0.6) & 0.6 & 0.5 (0.6) & 0.7 (0.7) & \textbf{0.3} (0.3) & 0.8 \\ \hline
\end{tabular}
\end{table}\section{Conclusion}\label{sec:conclusion}

For machine learning applications where minimizing RMS is a key factor, our DAE with metamorphic truth and imputation feedback outperforms both MICE and Gondara's  DAE imputation model. For removing bias from covariance relationships, our DAE surpasses Gondara's DAE and approaches and even exceeds MICE performance.
Additionally, the technique of metamorphic truth can be used to shape the way DAE or other neural networks learn. For example it may be possible to develop a metamorphism to guide the imputation to have better statistical characteristics, possibly by incorporating a regression or other MICE strategy into the metamorphism. Beyond the case of imputation, metamorphic truth could have wider applications in areas such as classification, that have yet to be explored.

%%
%% The next two lines define the bibliography style to be used, and
%% the bibliography file.
%\bibliographystyle{ACM-Reference-Format}
\bibliographystyle{acm}
\bibliography{our_bib}
\end{document}